\documentclass[runningheads]{llncs}

\usepackage[utf8]{inputenc}
\usepackage{graphicx}

\usepackage{float}
\usepackage{multirow}  % For table
\usepackage{amssymb}
\usepackage{amsmath}
\usepackage{fancyhdr}            % Permits header customization. See header section below.
\usepackage[colorlinks=true, citecolor=blue,linkcolor=blue,bookmarks=false]{hyperref}
\usepackage[labelsep=quad,indention=10pt]{subfig}
\title{Extreme Volatility Prediction in Stock Market: 
When GameStop meets Long Short-Term Memory Networks
}
\author{Yigit Alparslan
\and Edward Kim}

\date{February 2021}

\institute{Drexel University, Philadelphia PA 19104, USA \\
\email{\{ya332, ek826\}@drexel.edu}}
\titlerunning{}

\begin{document}

\maketitle

\begin{abstract}
The beginning of 2021 saw a  surge in volatility for certain stocks such as GameStop company stock (Ticker GME under NYSE). GameStop stock increased around 10 fold from its decade-long average to its peak at \$485. In this paper, we hypothesize a buy-and-hold strategy can be outperformed in the presence of extreme volatility by predicting and trading consolidation breakouts. We investigate GME stock for its volatility and compare it to SPY as a benchmark (since it is a less volatile ETF fund) from February 2002 to February 2021. For strategy 1, we develop a Long Short-term Memory (LSTM) Neural Network to predict stock prices recurrently with a very short look ahead period in the presence of extreme volatility. For our strategy 2, we develop an LSTM autoencoder network specifically designed to trade only on consolidation breakouts after predicting anomalies in the stock price. When back-tested in our simulations, our strategy 1 executes 863 trades for SPY and 452 trades for GME. Our strategy 2 executes 931 trades for SPY and 325 trades for GME. We compare both strategies to buying and holding one single share for the period that we picked as a benchmark. In our simulations, SPY returns \$281.160 from buying and holding one single share, \$110.29 from strategy 1 with 53.5\% success rate and \$4.34 from strategy 2 with 57.6\% success rate. GME returns \$45.63 from buying and holding one single share, \$69.046 from strategy 1 with 47.12\% success rate and \$2.10 from strategy 2 with 48\% success rate. Overall, buying and holding outperforms all deep-learning assisted prediction models in our study except for when the LSTM-based prediction model (strategy 1) is applied to GME. We hope that our study sheds more light into the field of extreme volatility predictions based on LSTMs to outperform buying and holding strategy.
\end{abstract}

\let\thefootnote\relax\footnote{{All code is open-sourced on \href{https://github.com/drexelai/gamestop-meets-lstms/}{GitHub.}}}

\section{Introduction}
\label{sec:introduction}

Stock market prediction has been an interesting study topic to many researchers \cite{ensemblestockpredictor}.
In recent years, the community also saw a surge in using deep learning assisted models to predicting the price of securities \cite{lstmpredictions}. The stock market price prediction via deep learning assisted models usually includes developing advanced mathematical models to learn patterns from sequential data with the goal of recognizing them later. Predicting future stock prices can help hedge-fund managers, retail investors, and institutions increase their portfolio returns  and can help data scientists and market researchers improve their mathematical models and transfer their learning from the domain of stock price predictions to other domains such as time series prediction, crime rate prediction, fair market value of goods predictions, etc. \cite{crimemachinepaper}.

On January 25th, 2021, the GameStop company stock (Ticker GME) hit an all-time high at \$485 and was trading about 10x of the decade-long average. On January 25th, the volume was 558 million and on January 4th, 2021, the trading volume was 33.63 million.  The extreme volatility presented many trading opportunities for the retail and institutional investors, even though the risks were high. 

In this paper, we hypothesize that a buy-and-hold strategy can be outperformed in the presence of extreme volatility by predicting and trading consolidation breakouts.
We investigate GME stock for its volatility and compare it to SPY as a benchmark from February, 2002 to February 2021.

We hope that our study sheds more light into the field of extreme volatility predictions based on LSTMs to outperform a buying and holding strategy.

This research is organized so that \autoref{sec:introduction} introduces the concept of stock predictions based on LSTMs and \autoref{sec:relatedwork} explores what has been done in this field. \autoref{sec:methodology} explains the dataset and the models in this study. We report the results in \autoref{sec:results} and conclude the study in \autoref{sec:conclusion} and \autoref{sec:futurework} with summarizing what we have done in this study and discussing where the research might go in the future.

\section{Related Work}
\label{sec:relatedwork}

Farmer et al. \cite{Farmer2254} studied the existence of randomness in the London Stock Exchange found out that randomly placed bets could explain 96\% of the variance of an individual stock price spread (the difference between the buy and ask price of a stock). Studies like Farmer are inspired by the following question: To what extent does the past stock price help predict the future stock price? Or in other words, does historical stock price contain rich information to predict future stock prices? These questions have been intriguing to many researchers since 1960s including that of Fama et al. \cite{fama1965behavior}. In recent years, researchers  studied extrapolating from the time series data by developing mathematical models \cite{tsay2005analysis}. Support Vector machines for time series have been studied \cite{svmpredictors} as well as Long Short-Term Memory Networks \cite{dnnstockpriceprediction}.

Ma et al. studied volatility as an independent variable of an universal distribution by looking at the binary strings of trading days and creating a distribution out of their kolmogorov complexities from the binary strings \cite{volatilitystudymaetal}. Their study on binary strings had a focus on learning from the structures present in stock's historical data and had the assumption that volatility can be studied independently from the financial information a stock might have in their historical data. We are inspired by such assumptions since in this paper, we treat the stock price prediction as a time series prediction problem independent from human psychology, market sentiment and data from financial indicators analysts use to make market predictions.

\section{Methodology}
\label{sec:methodology}

\subsection{Data}
We study SPY and GME stock adjusted close market close prices from February 13th, 2002 to February 12nd, 2021. We treat SPY exchange-traded fund (ETF) as a benchmark study since it is tracking S\&P 500 index and is less volatile compared to GME. We choose GME because of its recent increase in its trading volume and volatility in the beginning of 2021. 

\subsection{LSTM Architecture, Training, Testing and Data Preparation}
In this section, we give more information regarding training and testing the LSTMs for Strategy \#1 and \#2. Strategy \#0 is simply buy-and-hold strategy so we skip it here.

\subsubsection{Strategy \#1: Architecture}

We first split the dataset into training and testing set. We split the stock price data so that 80\% of data frame is for training and remaining 20\% is for testing. We also standardize stock prices to numbers between zero and one.

The input data to an LSTM model is a 2-dimensional array. The shape of the array is samples x look-back.

samples: the number of data points.
look-back: At time $t$, the LSTM will process data up to (t-look-back) to make a prediction. Very large look-back values usually result in accumulating error and very small look-back values usually result in failing to learn from data. We pick 60 as our look-back value. 

We convert the data into a 2D dimension array like this:
From the start of the dataset till the end, we extract 
windows of size 60. In the end, we end up with $X_train-60$
rows and 60 columns in our dataset.
 
Additionally, we add dropout layer to prevent overfitting.
When defining the dropout layers, we specify 0.2, meaning
that 20\% neuron activation will be set to 0. Next, we fit
the model to run on 100 epochs with a batch size of 32. 
 
Once training is done, we use the testing data set to measure our model's performance. In order to predict future stock prices we need to do as follows:
\begin{enumerate}
    \item Load the test dataset
    \item Standardize stock prices to numbers between zero and one
    \item Extract windows of size 60 from test dataset and stack them. After this step, each row has 60 values (60 is our lookback value)
    \item Make predictions
    \item After making the predictions, we use $inverse_transform$ to get back the stock prices in normal readable format.
\end{enumerate}

\subsubsection{Strategy \#2: Architecture}
We first split the dataset into training and testing set. We split the stock price data so that 80\% of data frame is for training and remaining 20\% is for testing. We also standardize stock prices to numbers between zero and one.

Anomaly detection approach we use in this paper is as follows:
\begin{enumerate}
    \item We train LSTM on data with no anomalies
    \item We then take data new data and try to reconstruct that using an autoencoder
    \item If the reconstruction error for the new dataset is above some threshold, we label that data point as an anomaly.
\end{enumerate}

The input data to an LSTM model is a 2-dimensional array. The shape of the array is samples x look-back.

samples: the number of data points.
look-back: At time $t$, the LSTM will process data up to (t-look-back) to make a prediction. Very large look-back values usually result in accumulating error and very small look-back values usually result in failing to learn from data. We pick 30 as our look-back value.

We pick threshold as 0.55, meaning that when we construct our input and get an mean absolute error of 0.55 or more, we label the input as anomaly. This threshold value is always between 0 and 1 because one benefit of standardizing data is to have a mean absolute error between 0 and 1.

\subsection{Trading and Backtesting Methods}

For our experiments, shorting a stock is never allowed. Additionally, only instruments that are allowed to trade are stock share securities. Option contracts are not allowed to trade for SPY and GME. During our backtesting, we can only sell a stock share that we have before we buy. Additionally, we have to sell our stock before we buy another share on the next day. So, we possess one GME share and/or one SPY share at any given time to simulate what day traders and swing traders do in real life with a limited budget.

\subsubsection{Strategy \#0: Buy and Hold Strategy.}
\label{sec:strategy1}
The Buy and Hold strategy is used as a benchmark to for our subsequent strategies. Our buy and hold strategy represents buying one single share at the first date of the time period we studied on the market close and selling the share at the market close on the last date. We report the profits as dollar (\$) amounts.

\subsubsection{Strategy \#1: Trading on Pure LSTM-based predictions.}
\label{sec:strategy2}
An LSTM architecture is used with a 60-day look-back period on a time series data that spans 19 years. Such short look-back period is chosen because LSTM are known to multiply the prediction error when predicting time series data far in the future. We backtest our strategy as follows:

\begin{enumerate}
    \item We split our stock data into train and testing.
    \item We train our LSTM based on train data stock movement for SPY and GME and predict the stock price for our testing data.
    \item Once we have the future predicted stock prices, for each day, we either buy a share to sell later to increase our profits, or wait. 
    \item We sell a share at its peak to capture maximum profit before the price drops.
    \item If the actual stock price on the test data set was not higher than the price we bought in, we close the trade with loss. If the actual stock price on the test data set was higher than the price we bought in, we close the trade with a profit.
    \item We report the total number of profitable and unprofitable trades as well as total profit and success rate. 
\end{enumerate}

\subsubsection{Strategy \#2: Trading based on consolidation breakouts.}
\label{sec:strategy3}
A consolidation is a pattern used by many day traders and usually defined as when a security price goes sideways for a given period of time. A consolidation breakout means a sudden price action for a given security after a period of consolidation. A breakout could be upwards or downwards depending on testing the support or resistance levels. For this strategy, we always trade breakouts as upwards signals, (i.e if there is a breakout detected, we treat it as a signal to buy).
We backtest our strategy as follows:
\begin{enumerate}
    \item We split our stock data into train and testing.
    \item We train our LSTM AutoEncoder based on train data stock movement for SPY and GME and predict breakouts for our testing data.
    \item We predict a breakout by detecting if the stock price fluctuates beyond a certain threshold.
    \item Once we have the future breakouts  predicted, we buy a share at the close price of the breakout day and sell three days later at the market close price to capture profits. 
    \item We pick three days as a hyperparameter for our backtesting strategy and we encourage readers to reproduce our results with different values for this hyperparameter.
    \item If the actual stock price on the test data set was not higher than the price we bought in, we close the trade with loss. If the actual stock price on the test data set was higher than the price we bought in, we close the trade with profits.
    \item We report the total number of profitable and unprofitable trades as well as total profit and success rate. 
\end{enumerate}

\section{Experiment Results and Evaluation}
\label{sec:results}
In \autoref{table:all_results}, we report all of our results for three of our strategies. Success rate for a strategy is defined as the ratio of profitable trades executed to all trades for that strategy. In our simulations, SPY returns \$281.160 from buying and holding strategy \autoref{sec:strategy1}, \$110.29 from strategy 2 with 53.5\% success rate \autoref{sec:strategy2} and \$4.34 from strategy 3 with 57.6\% success rate \autoref{sec:strategy3}. GME returns \$45.63 from buying and holding one single share, \$69.046 from strategy 1 with 47.12\% success rate and \$2.10 from strategy 2 with 48\% accuracy. Overall, buying and holding outperforms all deep-learning assisted prediction models in our study except for when LSTM-based prediction model (strategy 1) is applied to GME.

\begin{table*}[htpb!]
\centering
\caption{Dollar amount returns for each strategy when back tested for the period of February 2002, and February 2021. We also report profitable and unprofitable trade counts and their sum. Success rate  for a strategy is defined as the ratio of profitable trades to all trades executed for that stock. Buy-and-Hold strategy is referred as Strategy 0 and does not need calculating its profitable/unprofitable trade count since it represents one transaction. }
\vspace{1em}
\begin{tabular}{cc|c|c|c|c|c|}
\hline
\multicolumn{1}{|c|}{\textbf{Stock}} &
  \textbf{Strategy} &\textbf{Profitable} & \textbf{Unprofitable} & \textbf{Total} & \textbf{Success Rate} & \textbf{Profits (\$)} \\ \hline

% SPY
\multicolumn{1}{|c|}{\multirow{3}{*}{\textbf{SPY}}} &
\textbf{ Buy \& Hold } & N/A & N/A & N/A & N/A & \textbf{281.16}\\
\multicolumn{1}{|c|}{} & \textbf{ Strategy 1 } & 462 & 401 & 863 & 53.53\% & 110.29\\
\multicolumn{1}{|c|}{} & \textbf{ Strategy 2 } & 536 & 395 & 931 & 57.57\% & 4.34\\
\hline

% GME
\multicolumn{1}{|c|}{\multirow{3}{*}{\textbf{GME}}} &
\textbf{ Buy \& Hold } &  N/A &  N/A &  N/A &  N/A & 45.63\\
\multicolumn{1}{|c|}{} & \textbf{ Strategy 1 } & 213 & 239 &  452 & 47.12\% & \textbf{69.04}\\
\multicolumn{1}{|c|}{} & \textbf{ Strategy 2 } & 156 & 169 & 325 & 48.0\% & 2.10\\

\hline
\end{tabular}
\label{table:all_results}
\end{table*}

\section{Conclusion}
\label{sec:conclusion}
In this paper, we investigate the field of predicting stock price action in the presence of extreme volatility. 
We investigate GME stock for its volatility and compare it to SPY ETF as a benchmark (since it is a less volatile index fund) from February, 2002 to February, 2021. 
Overall, buying and holding outperforms all deep-learning assisted prediction models in our study except for when LSTM-based prediction model (strategy 1) is applied to GME.
We hope that our study sheds more light into the field of extreme volatility predictions based on LSTMs to outperform buying and holding strategy.

\section{Future Work}
\label{sec:futurework}
In this paper, we only looked at selling shares that we have during our backtesting studies. In other words, shorting was not allowed when we were backtesting for the time period that we picked. We hope to expand this approach to include shorting to see its effect on LSTMs based trading strategies. Additionally, in this paper, we limited the backtesting simulations to only trading stock shares and not option contracts. However, there exist option contracts, which are security instruments designed to capture profit when the stock price moves very large amount (either up or down) from a predicted stock price. In the future, we can look into including selling and buying option contracts
when detecting consolidation breakouts. Including option contracts to our backtesting studies in the future might be worthwhile to investigate in the search of outperforming buy \& hold strategy.

\newpage
\bibliographystyle{splncs04}
\bibliography{references}

\section{Appendix}
\begin{figure}[tb!]%
\setkeys{Gin}{width=120mm}
\centering
    \subfloat[]{\includegraphics{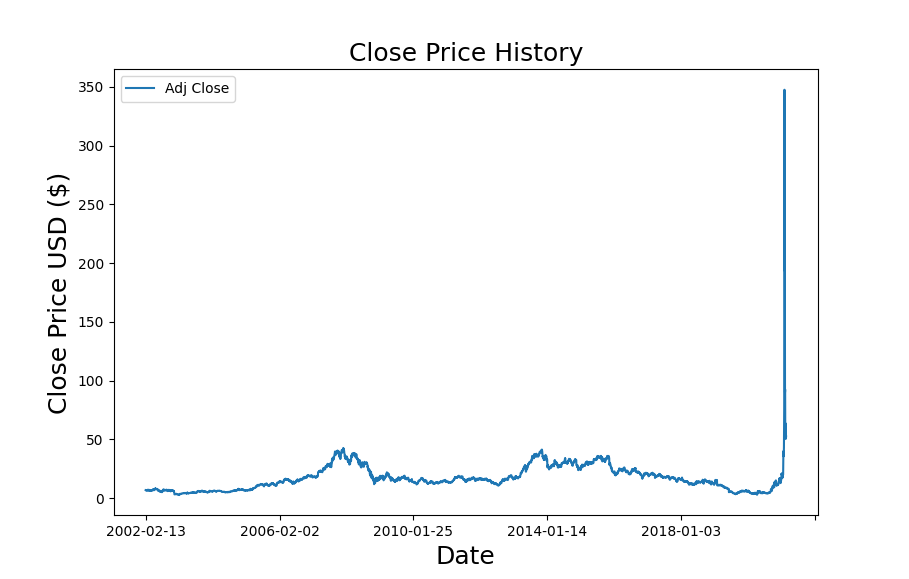} }%
      \qquad
    \subfloat[]{\includegraphics{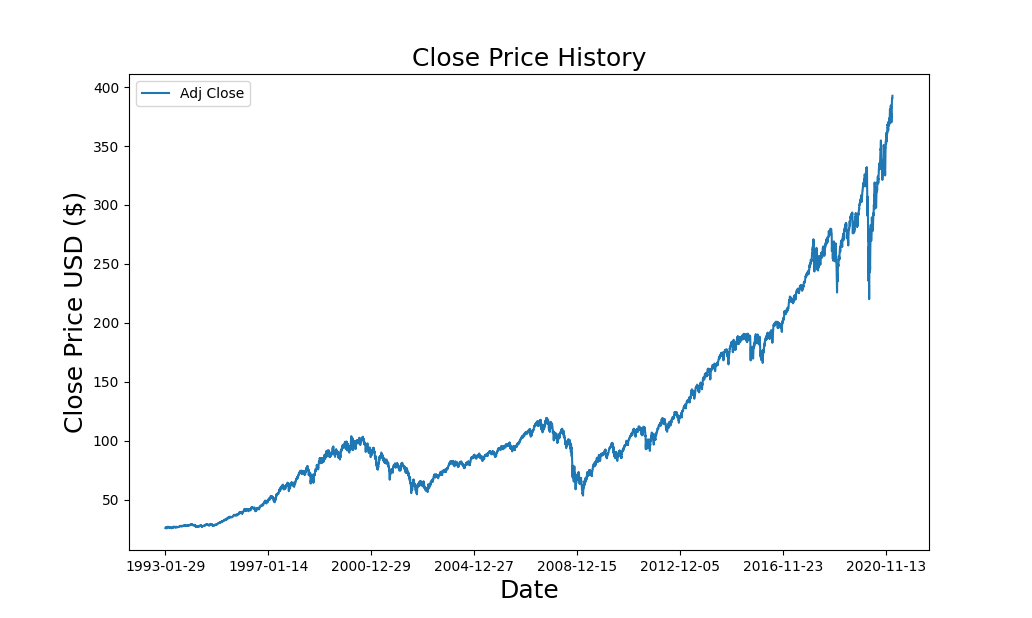} }%
    \caption{Comparison of (a) GME stock price at close from February, 2002 to February 2021 and (b) SPY stock price at close from February, 2002 to February 2021. The GME company stock has seen a surge in the price in the beginning of 2021 (about 10 fold increase).
 The close prices have been adjusted to negate the effect of dividends and splits.
}% 
\label{fig:stock_prices}
\end{figure}

\begin{figure}[tb]%
\setkeys{Gin}{width=100mm}
\centering
    \subfloat[]{\includegraphics{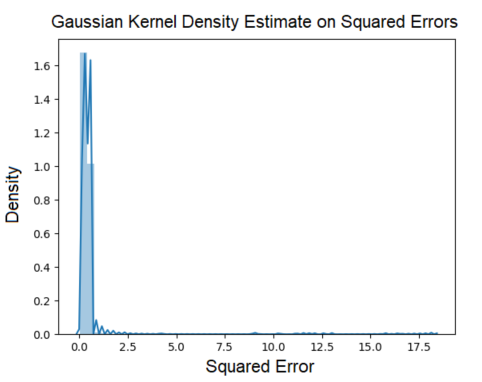} }%
    \qquad
    \subfloat[]{\includegraphics{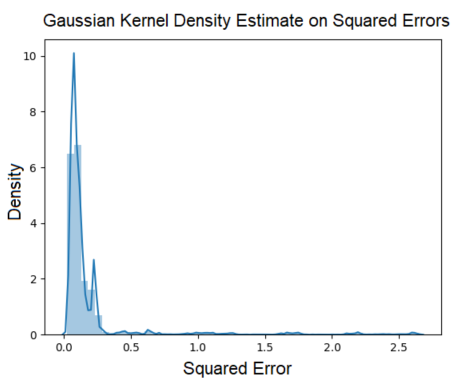} }%
    \caption{Comparison of (a) GME stock Gaussian density estimates on squared errors  when predicted on test data and (b) SPY stock Gaussian density estimates on squared errors  when predicted on test data. Each bin represents 50 predictions. 
}%
    \label{fig:train_errors}
\end{figure}

\begin{figure}[tb]%
\setkeys{Gin}{width=120mm}
\centering
    \subfloat[]{\includegraphics{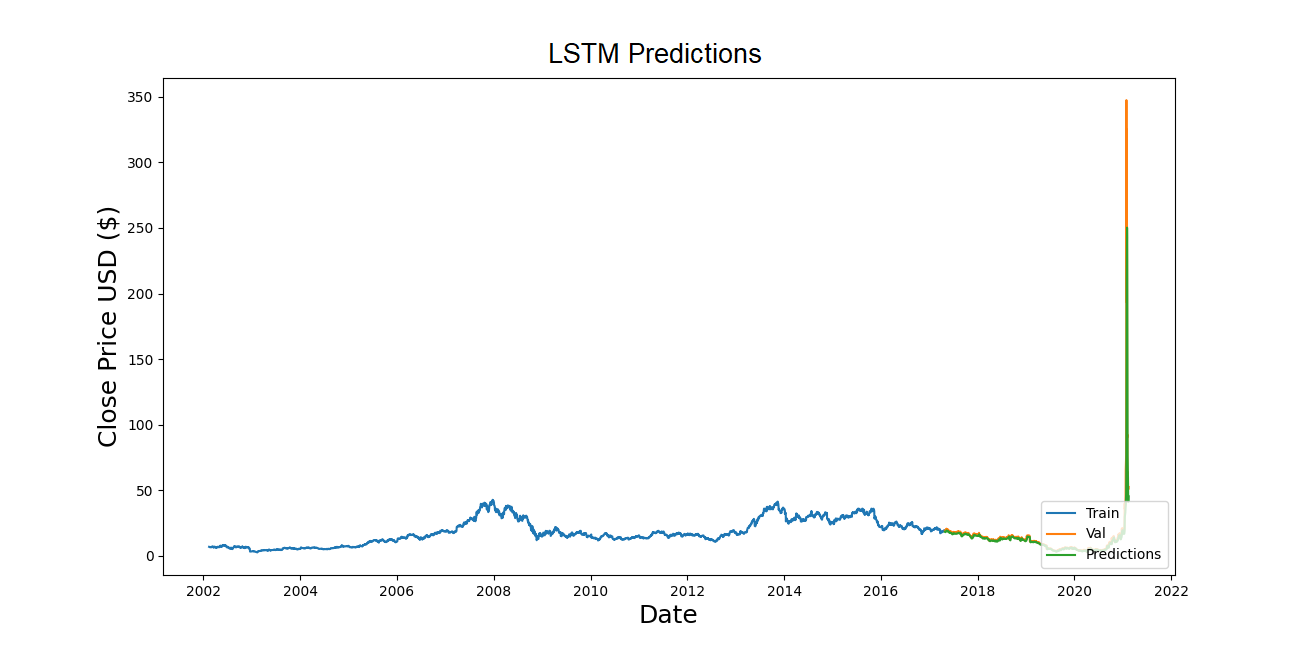} }%
      \qquad
    \subfloat[]{\includegraphics{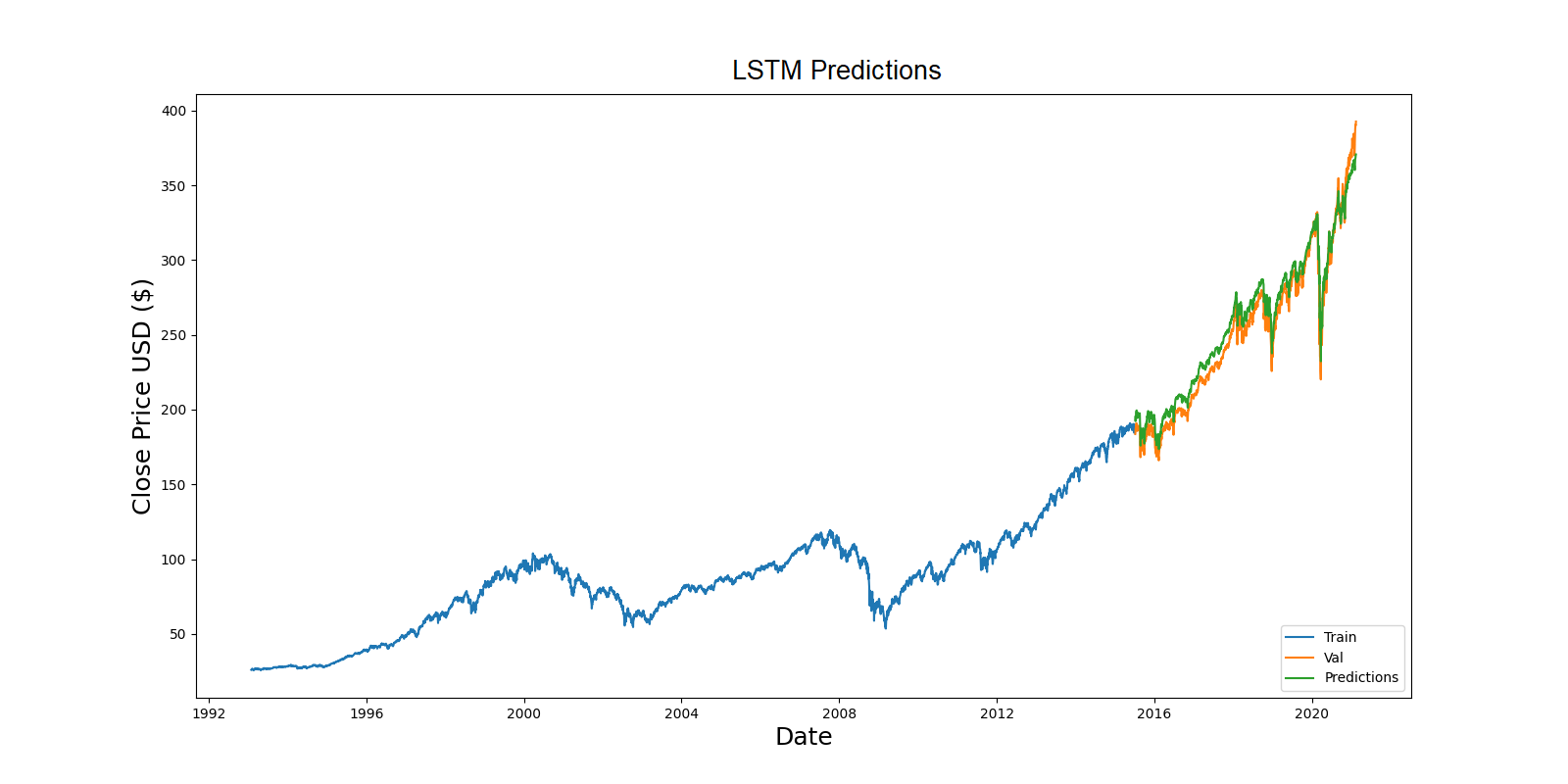} }%
    \caption{Comparison of (a) GME stock price prediction and (b) SPY stock price prediction. GME predictions return \$69.046 on 213 profitable trade vs 239 unprofitable trades and achieves success rate of 47.12\%. SPY predictions return \$110.29 on 462 profitable trade vs 401 unprofitable trades and achieves success rate of 53.53\%. 
}%
    \label{fig:predictions_strategy2}
\end{figure}

\begin{figure}[tb!]%
\setkeys{Gin}{width=120mm}
\centering
    \subfloat[]{\includegraphics{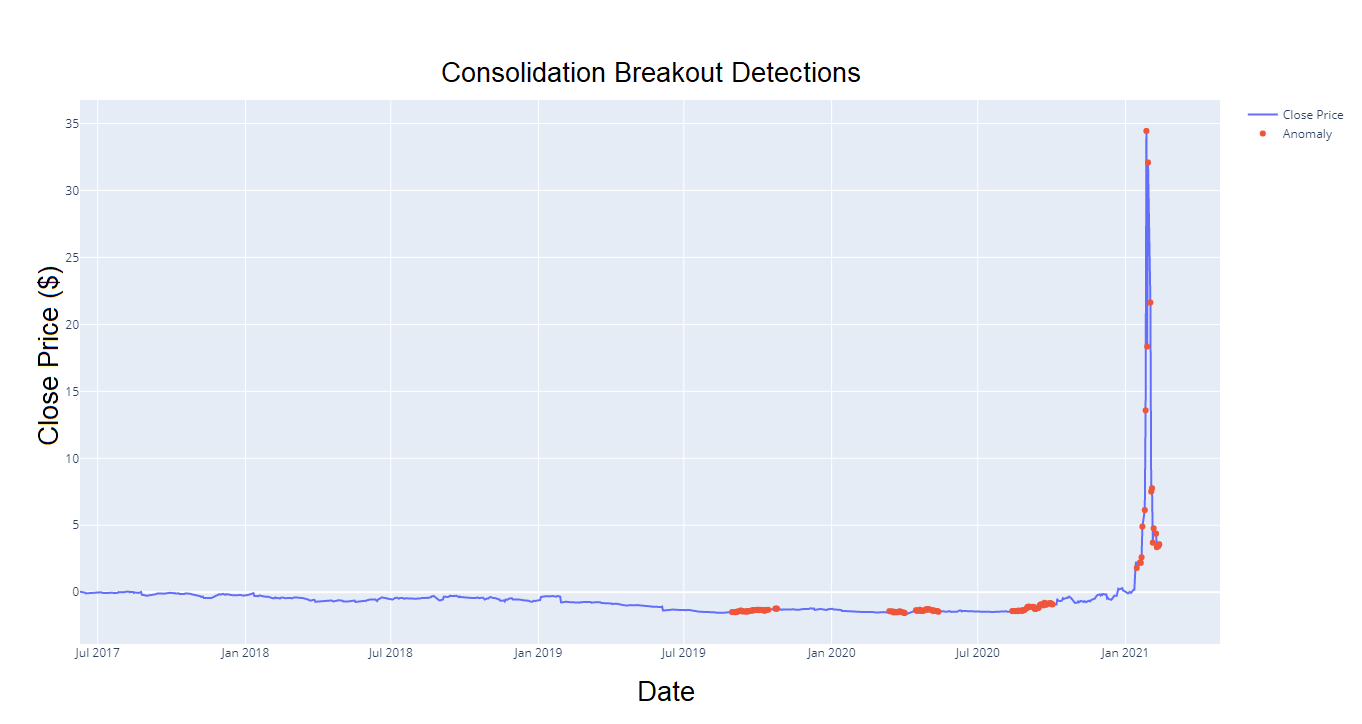} }%
      \qquad
    \subfloat[]{\includegraphics{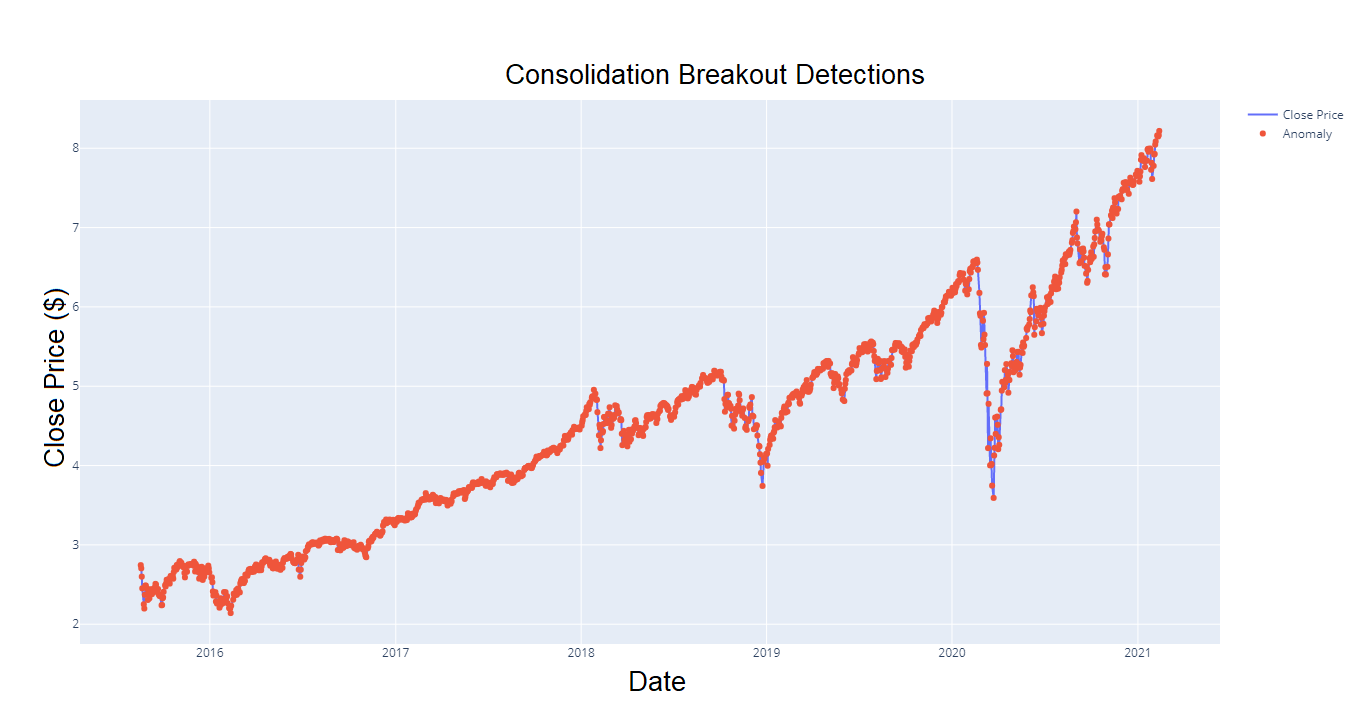} }%
    \caption{Comparison of (a) GME breakout predictions and (b) SPY breakout predictions. GME predictions return \$4.37 on 536 profitable trade vs 395 unprofitable trades and achieves success rate of 57.5\%. SPY predictions return \$110.29 on 156 profitable trade vs 169 unprofitable trades and achieves success rate of 48\%. 
}%
    \label{fig:breakouts_strategy3}
\end{figure}

\end{document}